\documentclass[12pt,oneside]{CUNY_CS_PhD}
\title{A Survey of Hierarchy Identification in Social Networks}
\author{Denys Katerenchuk}

\usepackage{fullpage}
\usepackage{graphicx}
\usepackage{amssymb}
\usepackage{url}
\usepackage{epstopdf}
\usepackage{enumitem}
\usepackage{microtype}
\graphicspath{ {Images/} }
\usepackage{todonotes}

\begin{document}

\frontmatter

\maketitle 
\makeabstractpage{Pr. Rivka Levitan}{  \textbf{Introduction:}
Humans are social by nature. Throughout history, people have formed communities and built relationships. Most relationships with coworkers, friends, and family are developed during face-to-face interactions. These relationships are established through explicit means of communications such as words and implicit such as intonation, body language, etc. By analyzing human interactions we can derive information about the relationships and influence among conversation participants.  However, with the development of the Internet, people started to communicate through text in online social networks. Interestingly, they brought their communicational habits to the Internet. Many social network users form relationships with each other and establish communities with leaders and followers. Recognizing these hierarchical relationships is an important task because it will help to understand social networks and predict future trends, improve recommendations, better target advertisement, and improve national security by identifying leaders of anonymous terror groups. In this work, I provide an overview of current research in this area and present the state-of-the-art approaches to deal with the problem of identifying hierarchical relationships in social networks.
}

\tableofcontents

\mainmatter

\chapter{Background}

\section{Motivation}
Imagine you are walking into a room full of strangers during your first job interview. You don't know who is who. Your task becomes to identify who is your future manager and who are your prospective colleagues. You have only one try to make the right impression on everyone. On one hand, there is a chance to start a casual chit-chat with the prospective manager and be perceived as not a serious employee, on the other hand, an overly formal tone of a conversation with a prospective colleague will make you look like a snob. Not surprisingly, people are pretty good at reading such situations and identifying who is who. One reason for it is that we make our guess on multiple sources of information. In particular, we look at the room and observe spatial relationships, the body language of each person, listen to the words and to the tone of the voice. All this information makes identification of a high-status individual a trivial task. Unfortunately, when analyzing online communities, all this information is unavailable and the problem becomes much harder.

Understanding hierarchical relationships in online communities is a crucial problem. For many people, online social networks (OSN) have become a major part of their social life. We share our happy life events, discuss and argue about the differences in our views, and meet new friends. However, the online world has its negative side too. For example, many terror and hatred groups have their online communities where they discuss malevolent topics while hiding behind anonymous identities. Following and understanding these discussions is an important step in preventing crimes. However, the discussions are often incomplete. If we can learn each user's stance on the subject and identify who is the community leader, we can predict the direction of the given conversation. In other words, knowing that a high-status user against some malicious plans, we can predict that this user will likely to convince the community users against the intention, and the opposite scenario is also true. Knowing the hierarchical relationships in OSN can prevent harmful events and even save lives.

Hierarchy prediction in online communities is a difficult problem despite the fact that scientists have been studying it for a long time (Section \ref{hist}). Throughout the study, a number of prominent corpora and evaluation measures have emerged (Section \ref{eval}). The majority of the data comes from websites such as Reddit and Twitter. Predicting user hierarchy from this data is challenging because the only signal of online communication is text. When a dataset contains a large number of users and interactions, one natural way to represent the network is a graph. For this reason, structure analysis algorithms are applied to this problem (Section \ref{sna}). The study of small datasets or datasets with partial information relies on text analysis (Section \ref{nlp}). Text analysis is mostly based on statistical methods proposed by linguists. Recently, with the advances in machine learning, neural network based methods have been applied in this domain (Section \ref{dl}). Despite all this work, there are still gaps in this research that should be addressed in the future work (Section \ref{fw}). We summarize the state-of-the-art research in this paper and conclude with our thoughts (Section \ref{conc}).   

\section{History}
\label{hist}
From the early history of the human kind, we have been trying to analyze social relationships. Many early philosophers, such as Plato, were concerned with the state of relationships and personal conduct. Plato's book The Republic presents discussions on the meaning of just behavior towards others. This book is arguably among the most influential pieces of literature and a point of reference for generations. Psychologists first defined the study of human science in 1669  \cite{gale1court}. T. Gale formalized the study to be a separate field from the divine science, which were indistinct at the time. This was the turning point, after which new research focused on studying face-to-face interactions and analysis of language, voice, and gestures. The scientists show that all these factors shape social connections and influence how people perceive each other \cite{ervin1984language, cappella1981mutual}. As the field expanded and involved larger population samples, analyzing thousands of individuals, new methods of analysis were needed.

\begin{figure}[h]

  \centering
    \includegraphics[width=0.35\textwidth]{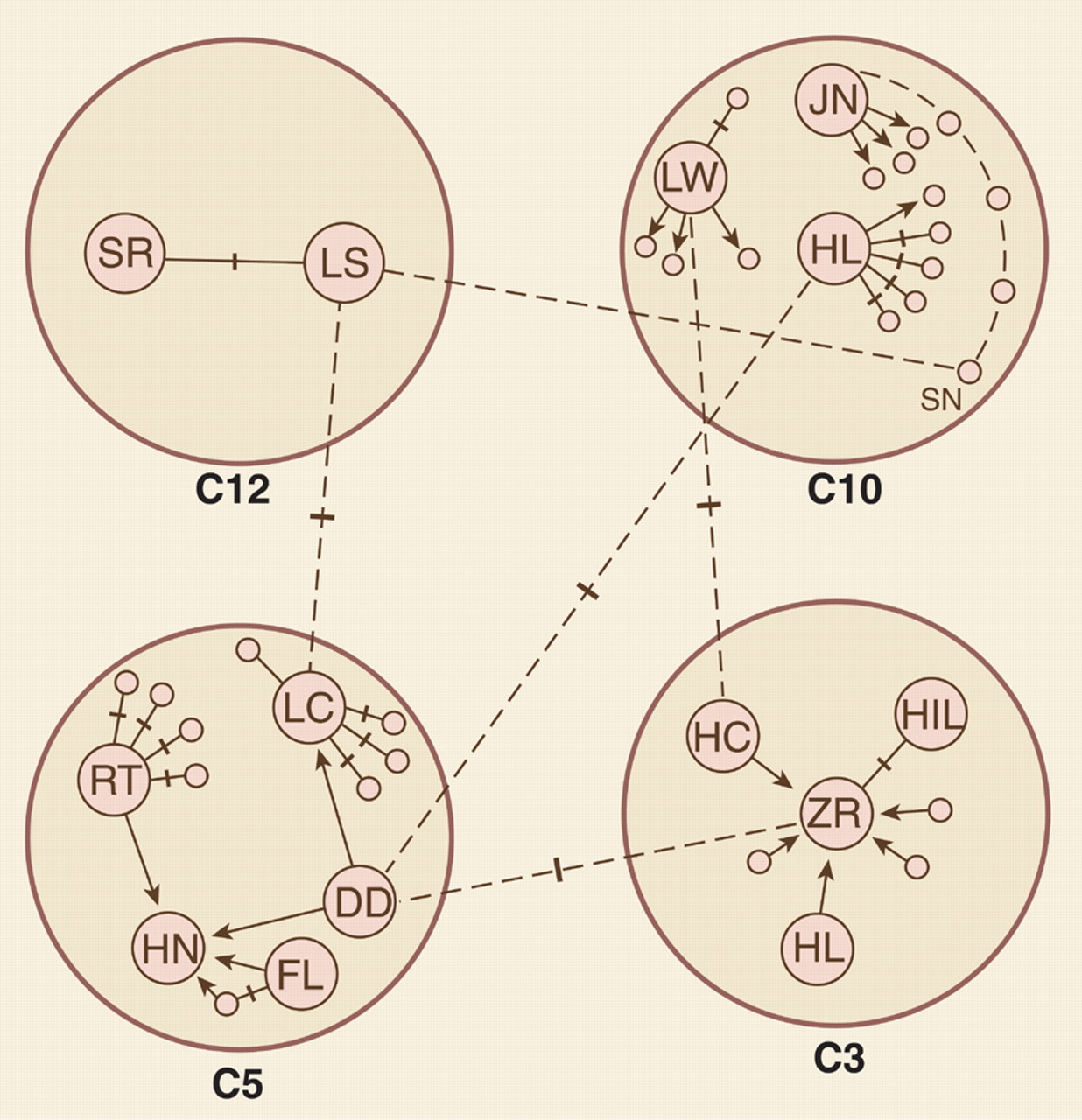}
   \caption{Moreno's social network of runaway children. The big circles represent residencies and the circles with letters represent runaway girls. (Image source: \cite{borgatti2009network})}
   \label{fig:moreno}
\end{figure}

A graph is a natural representation of interactions between people. Jacob Moreno applied network analysis in 1932 to investigate an unusual pattern of school run-away children \cite{moreno1932application}. Moreno noticed that the frequency of attempts was 30 times above the average among other schools and it was crucial to find the reason. His hypothesis was that the cause was social rather than environmental. When he plotted the relations among students as a graph (Figure: \ref{fig:moreno}), he noticed that the escapees were in the same social group and the escapes were results of internal trend \cite{borgatti2009network}. A. Bavelas formally defined mathematically the relations between the psychology of social groups and its topological structure \cite{bavelas48}. With the growing popularity of the Internet, a big part of human-to-human interactions has shifted online and social network analysis (SNA) became a fast-growing area of research. This, in turn, gave a start to a new domain of problems. The main objective of SNA is to understand relationships among the network participants. However, the structure analysis has one major drawback - it requires a somewhat complete network of interactions. Incomplete or dyadic communications don't have enough data to make a prediction. 
As a way to further improve the performance of SNA algorithms, scientists turned to study natural language.

Human language is a rich source of information. In online conversations, users communicate through text messages. The messages, besides the direct information, contain implicit information about the writer. James Pennebaker et al. \cite{stirman2001word} studied implicit information extraction on a problem of depression and suicidal author detection by looking at their writings. Their initial intuition was that authors with depression will be using more negative words. To their surprise, they found it noneffective.  While exploring other possible ways, they accidentally discovered that small and often not very informative words such as pronouns, articles, prepositions were indicative of the psychological state of a writer. This discovery gave a push to further exploration of the language. With the development of natural language processing (NLP) and machine learning (ML), text analysis has been a powerful tool. Current state-of-the-art algorithms are able to predict personal attributes such as gender, age, income, etc. \cite{Levi_2015_CVPR_Workshops,preoctiuc2015studying}. Identifying these attributes enables to understand social user interactions on the Internet. 

\section{Data}
\label{eval}
The area of social network analysis is very diverse with applications on a number of different online communities. These online communities, while inherently similar, serve a different purpose, hence, the objective is different as well. The most common differences are in the community type (emails, discussion forums, social networks, etc) and the conversation type (dyadic or multi-user). 

Among the variety or resources that has been in the center of research is the collection of emails from the infamous Enron corporation \cite{shetty2004enron}. The emails were collected during the investigation and later released to the public. This makes a great source of data for studying interactions between employees with defined status inside the company. The biggest advantage of this corpus is that the data comes from the real-world interactions and the hierarchy level is clearly defined by the job title.  In addition to the Enron corpus, websites such as Wikipedia \footnote{www.wikipedia.org}, StackExchange \footnote{www.stackexchange.com}, and Reddit \footnote{www.reddit.com} are used in research \cite{danescu2012echoes,danescu2013computational,zayats2017conversation}. Wikipedia is often used to understand dynamics of interactions on discussion pages where users propose edits to the articles. This creates a natural environment for conversations where user apply various language devices to make their point. The hierarchy levels are defined between editors and admins. StackExchange is a website where users post questions about their problems and the community helps to solve it. As a reward, the most active users earn points that correspond to status. Reddit website is a discussion platform where the users are free to post anything they desire and start the discussions. Reddit has an internal organization of topics, which are called subreddits. Similarly to StackExchange, Reddit has its own reward system in the form of karma score. The users can post comments and earn or lose karma points. This karma score is a proxy to status in a given subreddit. The status of each used is often separated into global - across a subreddit, or local - across a single thread-discussion. In addition to these resources, researchers often turn their attention to social networks.

Social networks have been a popular source of SNA research \cite{danescu2011mark,gilbert2009predicting}. Twitter \footnote{www.twitter.com} is a platform that allows posting short text message of 140 characters long (280 characters since September 2017) as well as pictures and videos. The users can broadcast tweets and can subscribe to follow each other. The number of followers as well as the number of likes is used as a proxy for influence. One interesting characteristic of Twitter is that accounts are public and the flow of messages is shown in somewhat linear fashion. This makes Twitter a great source of research data. Facebook \footnote{www.facebook.com} is another social network source that is used in research \cite{gilbert2009predicting}. Facebook user accounts are more private and the users often form networks of friends or form groups by interest. In order to join a network, a user would send a request. For this reason, Facebook friends are often friends or acquaintances in daily life. The users can post text and media to the network, but the posts are visible only to their connections. For this reason, data collection is challenging and requires user permissions. There are two Facebook corpora available: a corpus collected by Viswanath et al. in New Orleans region \cite{viswanath2009evolution} and a corpus collected by Wilson et al. \cite{wilson2012beyond}. Both corpora are similar with the first one focused on users that lives in a single region. 

One type of datasets that is only started to emerge recently is multi-modal social media data. These datasets link user profiles from different platforms. Farseev et al. \cite{farseev2017tweetfit} combined the data from Twitter, Instagram, Foursquare, and Endomondo. The goal of this work is to create a dataset to track user health with additional data on individual body mass index (BMI). However, this dataset can be used to study user behavior across different platforms. For example, this research can shed the light whether users that are influential in one community also have high status across the others. There is not much work done in this domain and further study is required.  

 \section{Evaluation}

Multiple rank-ordering measures exist to evaluate hierarchy prediction algorithms. In general, two common tasks are predominant in this area: 1) hierarchy identification in dyadic conversations \cite{Gilbert:2012:PSW:2145204.2145359}, 2) ranking all users according to their status level \cite{Agarwal:2012:CGS:2390665.2390706}. While the first task can be evaluated with a simple accuracy, the second task requires a measure that evaluates positions of each prediction with respect to the gold standard. The most common evaluations are surveyed below.


\subsection{Accuracy}
Accuracy is the common choice for classification type of problems. In general, accuracy is defined as a ratio of correct predictions and all possible predictions. 
 
\begin{center}
$Acc = \frac{t}{n}$\\
\end{center}
where t - true predictions and n - the number of samples.

Accuracy is a simple measure that works on multi-class datasets where the classes are balanced. When the class distribution is skewed towards one of the classes, the accuracy is high for a classifier that assigns the majority class to all instances. For this reason, other measures are a better choice in such cases.

\subsection{F-measure}

F-measure or F-score \cite{Rijsbergen:1979:IR:539927} is a common evaluation measure that is used to measure information retrieval algorithms. This measure is defined as follows:
\begin{center}
$$F = 2*\frac{p * r}{p + r},$$
\end{center}
where p - precision and r - recall.

Precision measures the portion of retrieved elements that are relevant. Recall measures the portion of relevant elements that were retrieved. This measure is a good choice for binary classification and not appropriate for ranking all users with multiple rank classes. 

\subsection{Average Precision and Mean Average Precision}

Average Precision (AP) \cite{zhuap} is a measure that is designed to evaluate information retrieval (IR) algorithms. AP works with unbalanced classes, where the number of elements of some class is dominant. AP measures precision at each element, multiplies the change in recall from the previous step and averages the results over the element list.  There exists a variation of AP that takes into consideration only the first k elements \cite{turpin2006user}, however, we will not focus on this variant. The formula to calculate the AP is the following:
\begin{center}
$$AP = \frac{1}{n} \sum\limits_{k=1}^n P(k) * \Delta R(k)$$
\end{center}
where $P(k)$ = precision@$k$ and $\Delta R(k) = |recall(k-1) - recall(k)|$.

Researchers often use mean average precision (MAP) \cite{liu_mean_2009}, which is defined as the mean of AP over multiple queries.
\begin{center}
 $$MAP = \frac{\sum\limits_{q\in Q} AP(q)}{|Q|},$$
\end{center}
where Q = a set of ordering problems and q = a single evaluation instance. 

Both AP and MAP measures have been designed to evaluate rank-ordering problems. The measures, however, assume no ties among ranks which manifests in inconsistent lower bounds.  Furthermore, these measures evaluate all rank values with equal cost. However, identification of very few high-rank items require more emphasis than over-represented low-rank items. This creates a problem where identification of many low-rank items produces a high score, despite the fact that these element of a lesser importance to the task. 

\subsection{Kendall's $\tau$}

Kendall's $\tau$ \cite{kendall38} is a correlation measure. This measure is often used when evaluating rank-ordering results. The measure considers the number of element pairs in reference and hypothesis lists and checks whether the relative orderings agree. The formal definition of Kendall's $\tau$ is shown below: 
\begin{center}
$$\tau = \frac{c - d}{\frac{1}{2} n(n-1)},$$ 
\end{center}
where c - a number of concordant (i.e. a correct relative ranking) pairs, d - a number of discordant (i.e. an incorrect relative ranking) pairs, and n - a number of pairs.\\

Kendall's $\tau$ is a popular choice for rank evaluation. Unfortunately, this measure also has some drawbacks. First of all, it does not explicitly deal with multiple ties and non-normal rank distribution. This will lead to a problem when an algorithm assigns the same (majority) rank value to all elements. Secondly, Kendall's $\tau$ does not produce a consistent lower bound score when the ranks follow a non-normal distribution. In addition, the score is produced by comparing the number of correlated elements and it does not emphasize rare high-rank elements. For these reasons, Kendall's $\tau$ is not the best choice to evaluate rank-ordering problems.  

\subsection{Discounted Cumulative Gain}

Among all evaluation measures, Discounted Cumulative Gain (DCG) \cite{Jarvelin:2002:CGE:582415.582418} has multiple advantageous characteristics to address a rank-ordering evaluation mentioned in the previous section. The main distinction of DCG from others measures is the ability to address non-normal rank distribution by assigning a higher cost to high-rank elements. This emphasizes the high-rank element identification. The formal definition of DCG is defined below:
\begin{center}
$$DCG = \sum_{i=1}^n \frac{rel(x_i)}{log_2(i+1)},$$\\
\end{center}
where n - a number of elements and rel() - some relevance function of the i-th element in a given list.

For comparison across multiple tasks, a normalized variant of DCG, nDCG, is calculated in the following way:

\begin{center}
$$nDCG=\frac{DCG}{IDCG},$$ 
\end{center}
where IDCG - represents the ideal DCG.

Unfortunately, this evaluation also has drawbacks. One drawback is this evaluation metric was designed for information retrieval rather than ordering evaluation. This means that one objective this measure considers is the number of relevant documents retrieved. Since all elements in the rank-ordering task are relevant, the measure's lower bound is never equal to zero. As a result, the range of prediction is from some arbitrary number between 0 and 1 to 1, which is the perfect score. This makes the comparison of different ordering problems hard hence there is no known lower bound. 
Another drawback is the cost function. While it addresses our concern to have different cost for high and low rank elements, we find in practical applications that both versions over estimate cost put on high rank elements. A more ``balanced" cost function would work better. Lastly, standard DCG produces different cost based on the element positioning. For example a list [9,1,1] will have different costs for [1,9,1] and [1,1,9]. However, we believe that a better way to consider the two lists as equally wrong.  Since the reference contains a tie in position 2 and 3, both of the hypothesized ranks are assigning the same rank to the element with relevance 9.  A useful way of visualizing this is as follows: since the reference list is [9,\{1,1\}], this requires treating the hypothesized lists as [1,\{9,1\}] and [1,\{1,9\}], where the relative position of the last two elements is irrelevant. To address these issues, we propose a new measure designed to evaluate ranking - RankDCG \cite{katerenchuk2016rankdcg}.

\chapter{Methods}
In the area of SNA two distinct methods exist to approach the problem: 1) structure analysis, and 2) language analysis. Each of these methods has its pros and cons. In this chapter we outline each of them and review the most prominent algorithms.

\section{Structure Analysis}
\label{sna}
A graph is a natural representation of a social network. Formally, a graph \textit{G} is a social network where a set of vertices \textit{V} represents the social network users and a set of edges \textit{E} represents connections between the users. The graph, depending on the social network, can be directed or undirected with a weight assigned to each connection. By using this representation the problem of hierarchy detection of each user can be solved by applying various graph centrality measures \cite{johnsen2017feasibility}.

\paragraph{Degree Centrality}
A degree centrality is the simplest measure that is based on the number of connections. It assigns a score by summing up all users connections. This measure assumes that the influential users will have more connections. 

\begin{center}
$$ C_D (\textit{v}) = \sum_{u\neq v, u\in V} \{1 | \textrm{ if (v, u)} \in E, 0 \textrm{ otherwise} \} $$
\end{center}

\begin{figure}[h]
\centering
\includegraphics[width=0.6\textwidth]{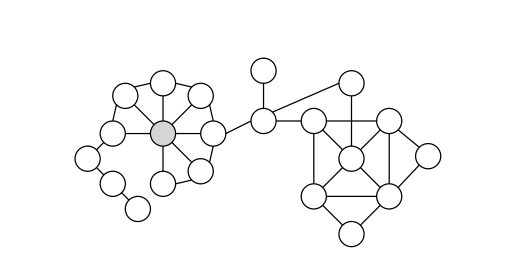}
\caption{Degree Centrality (Image source: \cite{johnsen2017feasibility}).}
\end{figure}

\paragraph{Closeness Centrality}
A closeness centrality is the measure of distance between a given node and every other node in the network. The intuition behind why this measure works for influence detection is that these users are closely connected with the entire community. 

\begin{center}
$$ C_C (\textit{v}) = \frac{1}{\sum_{u\neq v, u\in V} dist(v,u) } $$
\end{center}

Where dist(v,u) is a function of distance between v and u.

\begin{figure}[h]
\centering
\includegraphics[width=0.6\textwidth]{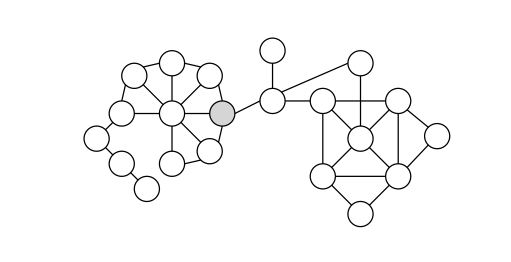}
\caption{Closeness Centrality (Image source: \cite{johnsen2017feasibility}).}
\end{figure}

\paragraph{Betweenness Centrality}
The measure hypothesis is based on the idea that the influential users often connect other users in the community and play a role of a bridge. The score is calculated by adding the number of times a user is between every other pair of users. 

\begin{center}
$$ C_B (\textit{v}) = \sum_{s\neq t\neq v, s\in V, t\in V} \frac{\sigma_{st}(v)}{\sigma_{st}} $$
\end{center}

where $\sigma_{st}$ - total number of paths from s to t, $\sigma_{st}(v)$ - number of paths through v.

\begin{figure}[h]
\centering
\includegraphics[width=0.6\textwidth]{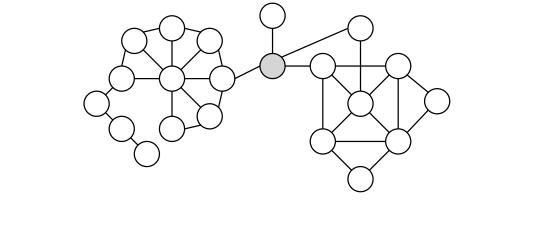}
\caption{Betweenes Centrality (Image source: \cite{johnsen2017feasibility}).}
\end{figure}

\paragraph{Eigenvector Centrality}
This centrality is based on the idea that important nodes are connected to other important nodes \cite{bonacich1972factoring}. In other words, users with high social status are friends with other high-status members and the reversed assumption is also true. The idea of this measure is similar to another popular algorithm, PageRank \cite{page1999pagerank}.   

\begin{center}
$$ C_E (\textit{v}) = \frac{1}{\lambda} \sum_{t \in M(v)} t = \frac{1}{\lambda} \sum_{t \in G} a_{v,t} t $$
\end{center}

where M(v) - a sat of neighbors of v, $\lambda$ - a constant, $a_{v,t}$ - is equals 1 if \textit{v} and \textit{t} have a link.
 
\begin{figure}[h]
\centering
\includegraphics[width=0.6\textwidth]{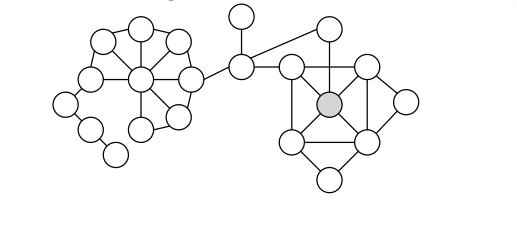}
\caption{Eigenvector Centrality (Image source: \cite{johnsen2017feasibility}).}
\end{figure}

These methods have shown to perform well on problems of identifying influential users. After the Enron crisis, many researchers took on the communication analysis during the investigation. Enron corporation had a well defined organizational structure from a regular employee to the CEO. This structure is used as the ground truth for hierarchy. Diesner et al \cite{diesner2005exploration} discovered that communicational patterns before and during the investigation were different. The employees were less active before the investigation with communication flow directed from the senior leaders to regular employees.
Closeness centrality was higher for top-ranked employees right before the investigation and for lower-rank workers during the investigation. This shows that before the scandal, Enron's culture was highly segmented with the directions sent from the top representatives. Eigenvector centrality was correlated with high-status employees the most. This is due to the formation of cliques inside the corporation. This work reveals the people responsible for the collapse were the top managers with tight cliques and internal top-to-bottom communicational structure.
   
The centrality measures are indicative of high-status users. \cite{Agarwal:2012:CGS:2390665.2390706} showed that degree centrality approaches outperform text analysis based methods proposed by E. Gilbert \cite{Gilbert:2012:PSW:2145204.2145359} (Section \ref{nlp}) for high-status user identification. This work is based on a larger Enron dataset with more employees. Having a large network is the main condition for graph-based algorithms to perform well. On the other hand, the eigenvector approach did not perfom well on this data. 

While Agarwal et al. claim that the graph-based structure analysis is superior to language-based analysis is valid, it also has a number of shortcomings. First of all, the results were tested only on a single corpus and it is necessary to compare the performance of both methods on a variety of datasets. Second, the eigenvector centrality measure did not perform well without any clear explanation. A deeper analysis into this would help to understand the problem. Third, the structure-based methods require access to the whole network and are unlikely to perform well on smaller datasets or dyadic conversations. In fact, J. W. Johnsen and K. Franke \cite{johnsen2017feasibility} showed that these centrality measures don't work well on loosely structured datasets. For these reasons, language-based analysis for identifying influential users cannot be ignored. 

\section{Language Analysis}
\label{nlp}
The word choice in a text can reveal a lot of information about the writer. Gender, native language, age, emotions and information can be derived from a sample of writing. This section introduces common practices and underlines the pros and cons of each approach. 


\subsection{N-grams}
N-grams are simple multi-word counts sorted by their frequency. The counts are stored in a vector that represents a document and each count position corresponds to a single word from the document. This representation is common throughout the area of NLP. One advantage of this method is that each document is represented by a point in an n-dimensional space where n is equal to the vocabulary size. Thinking of a collection of documents as a collection of points transforms the problem into finding a division boundary. This can be achieved with classification models such as SVM \cite{smola1997support} or random forest \cite{liaw2002classification}. Unfortunately, there are downsides to this approach. One major problem is the size of the vector. It can get huge considering that each position corresponds to a single word. For example, it is approximated that there are around half a million words in the English language\footnote{https://www.merriam-webster.com/help/faq-how-many-english-words}. Working with vectors of this magnitude becomes unfeasible. Another issue is that the vectors are sparse with most values equal to zero. In such scenario, most machine learning algorithms don't perform well. For this reason, only a limited subset of words is used. They are either manually selected based on the domain knowledge, limited to the most frequent words, or words that represent categories (emotion, self-promotion, achievement-related, etc.).

Document representation based of word-frequencies is a simple and effective approach. However, word categories and domain-expert selected words require manual labor. For this reason, using publicly available datasets is the optimal solution. WordNet \cite{fellbaum1998semantic} is a lexical database of English where each word is grouped according to its semantic relationships. The words "table" and "chair" are part of a larger group "furniture". Combining words into higher order groups allows for dense document representations with fewer dimensions.

James Pennebaker, discovered that stop words (articles, pronouns, prepositions, conjunctions, and auxiliary verbs) are markers of one's emotional and psychological state \cite{pennebaker2003psychological}. This discovery was crucial because these words were often ignored during text analysis \cite{bramsen2011extracting}. In later work, Pennebaker et al. developed a list of word categories (LIWC) that are associated with an author's cognitive state \cite{pennebaker2015developmentliwc}. The list contains categories such as work, time, feel, positive/negative emotions, etc. The categories are manually constructed based on psychological analysis of humans. This list serves two main purposes: 1) includes words that are known to be indicative of one of the classes, 2) reduces the size of document representation by mapping multiple words into a single class. All these techniques were successfully applied to detect the level of influence \cite{bramsen2011extracting, Gilbert:2012:PSW:2145204.2145359}.

Bramsen et al. \cite{bramsen2011extracting} successfully applied the n-gram based approach to the Enron corpus. The task was to predict whether an email was sent to a recipient of a higher status. This problem is defined as a binary classification of a single email. They achieved accuracy of 0.78\% with an SVM classifier. Eric Gilbert expanded on this method by including LIWC word list for identifying the most prominent phrases that signal workplace hierarchy \cite{Gilbert:2012:PSW:2145204.2145359}. This word list of phrases is an attempt to develop a resource similar to LIWC, but for hierarchy detection. During this work, he discovered that some phrases are, indeed, indicative of a high status. For example, phrases such as "let's discuss", "any comments", "we are in", etc. are signals that a sender has higher status than the recipient. However, the list also includes corpus specific phrases such as "Europe" or "worksheet". Without further cleaning, the word-list might be specific to a given dataset and minght not not generalize well to achieve the same results on other datasets. Nevertheless, n-gram and list-based approaches consistently show high performance without going into deep structural or linguistic analysis.

\subsection{Hedging}
The use of hedges in conversations signals social status. Lakoff et al. first introduced the term "hedge" in 1973 in his theoretical paper \cite{lakoff1973hedges}.  A hedge is a linguistic device that is used in conversations to mitigate the meaning of a statement, request, or question.  An example of a hedge phrase can be "Whenever you have some time let's give it a try". If you ever received such a request from your manager you know that it means you should do it now. The hedges are often used in a superior/subordinate kind of relationships to mitigate direct orders. Identifying hedge phrases can help to understand the relationships between speakers and find uncertain statements.  Hedge identification is a hard problem because hedges are regular words that are used in a slightly different context. 
The CoNLL-2010 shared task is one of the most common datasets that is concerned with hedge identification. The dataset is based on data from the Biomedical domain and Wikipedia discussion pages \cite{farkas2010conll}. Each sentence in the corpus is labeled as "certain" or "uncertain" if it contains a hedge phrase. In addition, the hedge phrases are annotated with the phrase boundaries for phrase identification. Complex verb phrases and passive dummy subject forms ("there is/are") were annotated as hedge cues as well.

A number of different methods have been proposed to find sentences that contain hedge phrases. Choi et al. \cite{choi2012hedge} based their work on simple n-gram model with human annotated hedge-phrases, which are motivated by domain knowledge. They were able to achieve F1 score of 0.65 on Biomedical data and F1 of 0.45 on Wikipedia beating the baseline defined in the CoNLL-2010 challenge. Despite the improvements, this model is quite simple. The biggest problem is that the trained model works well only on the specific domain. For example, a model trained on Wikipedia does not work on the Biomedical domain. Jean et al. \cite{jean2016uncertainty} introduced a probabilistic model that addresses this problem. The main idea is that each sentence is represented as a tuple of size six where each feature corresponds to a probability over some specific dimension. The dimensions are Lemma based uni-grams and bi-grams in certain and uncertain sentences, part-of-speech 5-grams, and max count of a Lemma for uncertain sentences. Formally, the probabilities are defined as follows:

\begin{center}
$F_i(s) = \sum_{k=1}^n p_i(c|w_k) \times conf(w_k),$
\end{center}

where \textit{s} is a sentence, \textit{c} is a probability of a class, and \textit{conf()} is a confidence function such as $conf(w) = 1 - \frac{1}{\#s(w)}.$

This model achieves F1 score of 0.57 on the Wikipedia data outperforming the work of \cite{choi2012hedge}. The big part of the improvement comes from the $conf()$ function. When this is removed, the model produces the F1 of 0.20. Despite the claim that this model should generalize better, the paper does not provide any cross corpora performance results. For this reason, model generalization remains one of the problems in identifying hedges.

\subsection{Entrainment}
Entrainment is a way collocutors mimic each others style of communication on the non-conscious level. Entrainment comes from the idea of linguistic style matching (LSM) and the change in style has been linked to correlation with social status.  \cite{ireland2011language} performed an analysis of speed dating transcripts and instant messaging conversations of couples. The paper showed that people mimic each other style of conversations by coordinating usage of function words. Formally, the relation is defined as follows:

\begin{center}
$LSM_{preps} = 1 - \frac{|preps_1 - preps_2|}{preps_1 + preps_2 + 0.0001}$
\end{center}

where $preps$ - preposition category from LIWC, $preps_1$ and $preps_2$ correspond to conversation participants.

This model of calculation is done using word counts that are contained in one of the LIWC categories. The rate of change specifies the entrainment. In this way, the low score indicates low language coordination. They discovered that the speed dating couples whose score was above the average, were  33\% more likely to meet in person. The couples whose score was below the average, only 9\% showed the desire to meet. From the analysis of married couple's instant messages, high language coordination score was correlated with long-term relationship stability. This early work revealed that entrainment has positive correlation with the level of attraction among conversation participants. While this work shows positive results, it also lacks the ability to capture the initial style of each conversation participant and how the style changes throughout the conversations. For example, it is not clear whether the collocutors were actually mimicking each others style or they just have a preference for someone who is similar to them. This question was addressed in a work done by Danescu-Niculescu-Mizil et al on Twitter \cite{danescu2011mark}. Twitter is a huge resource of data but each individual tweet is only 140 characters. For this reason, the conventional algorithms for entrainment that were designed to analyze long texts don't work and they needed to improve a way of capturing the change in style. Danescu-Niculescu-Mizil et al introduced a notion of personal background style and performed a temporal analysis. In other words, a user can entrain only after receiving a message by mimicking the style. This can be defined as follows: given a conversation between two users \textit{a} and \textit{b}, and the use of some stylistic dimension C, what is the change in the probability of user \textit{b} to use the same class C.

\begin{center}
$Acc_{a,b}(C)\triangleq P(T_b^C |T_a^C, T_b \hookrightarrow T_a) - P(T_b^C|T_b \hookrightarrow T_a$
\end{center}

where $T_a^C$ and $T_b^C$  are the events of users \textit{a} and \textit{b} use category C.

This framework is better suited to capture the entrainment in conversations. However, the results were inconclusive for a couple of reasons: 1) the dataset was very large, 2) the conversations were limited to only 140 characters, 3) the labels for social status weren't clearly defined (\# followers, \# followees, \# posts, \# days on Twitter, \# posts per day). From the analysis, only the rate of personal pronouns was correlated with the number of followers. Nonetheless, when the same approach was applied to different datasets, Danescu-Niculescu-Mizil et al. achieved successful results \cite{danescu2012echoes}. The experiment was done on Wikipedia talk pages and U.S Supreme Court transcripts. The choice of the datasets was determined by the need to address two different types of power: earned (Wikipedia) and situational (Supreme Court justices).
By counting a change in the usage of LIWC word categories \cite{pennebaker2015developmentliwc} they discovered a few interesting cases. First, they found that users on Wikipedia talk pages tend to coordinate their language more towards administrators, also referred to as admins. On the other hand, the admins entrain more than regular users in their language in general. This case is interesting because it shows that the admins might use their language coordination to influence the community and persuade their point of view. This supported by an analysis of admins' communication and revealed that to-be-admins coordinate their language more than regular users, but stop coordinating it as much when they reach a high position in the community. From the study of Supreme Court data, they found the opposite situation. In the court, lawyers, who have lower rank, coordinate their language more than justices. This fact suggests that lawyers might use their language to influence justices in their favor. While there is a difference between the two corpora, the results from both show that entrainment is suggestive of hierarchical relationships and that different kinds of power/hierarchy may need to be modeled differently


\subsection{Politeness}
\begin{figure}[h]

  \centering
    \includegraphics[width=0.6\textwidth]{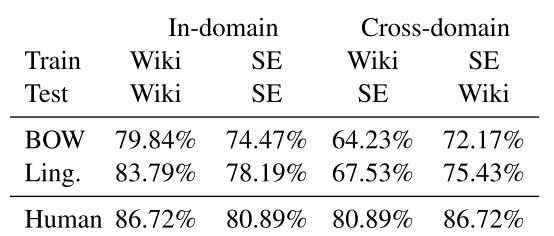}
   \caption{Classification accuracy for Wikipedia and StackExchange datasets}
   \label{politeness_results}

\end{figure}

Politeness in conversations has been linked to social power dynamics. Many theories propose that individuals with lower social status employ a polite tone of conversation while communicating with a higher status person \cite{lakoff1977you}. Danescu-Niculescu-Mizil et al. analyzed online social websites such as StackExchange and Wikipedia discussion pages \cite{danescu2013computational}. This work is based on the n-gram model as well as on linguistically informed words where humans define word-phrases and extract them with regular expressions. Each linguistically informed phrase is a part of a class such as "Gratitude" and an expression that matches "\textbf{I} really \textbf{appreciate} ..." or "Hedges" with "I \textbf{suggest} we ...". In total there are 20 linguistically informed word classes.
The prediction is done at leave-one-out cross-validation with SVM classifier. The results can be found below in the table \ref{politeness_results}.

The results show high accuracy for predicting which messages are polite and which are not. The interesting point is that the human-constructed phrases help to improve n-gram approach. In Wikipedia data, politeness was correlated with the high status users. On StackExchange website, users that use polite tone in their questions are more likely to get helped. In addition, the social status on StackExchange also correlates with politeness with a caveat: a user that is trying to establish the social position in the community tends to be more attentive and polite answering the questions. However, as soon as they reach high status, they stop being polite and become offensive. The politeness analysis can bring more information to understand relationships in text communications.

\section{Neural Networks}
\label{dl}
\begin{figure}[h]
\centering
\includegraphics[width=0.7\textwidth]{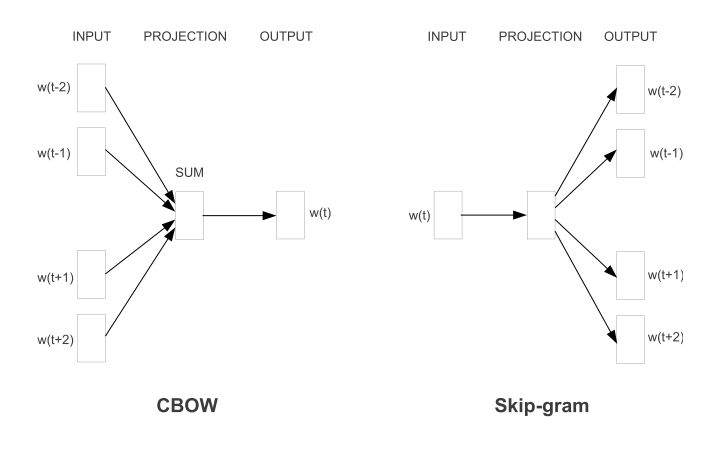}
\caption{CBOW and Skip-gram model architectures.}
\label{nn_lang}
\end{figure}

Neural Networks (NN) have become a prominent method to address multiple problems in the area of artificial intelligence including vision, robotics, and NLP. Text analysis is one example where NN consistently make improvements. In 2013, Mikolov et al. \cite{mikolov2013efficient} proposed a new technique to improve training of neural network language model that was first proposed by Hinton et al. \cite{Hinton:1986:DR:104279.104287} and Bengio et al. \cite{bengio2003neural}. The main idea comes from linguistic distributional semantic theory that states that the words that are similar in meaning will occur in a similar context. Each word is represented by a randomly initialized vector of a predefined size. Then, given a training data, the vectors are updated from the context they appear. Two approaches were defined for updating weights: continuous-bag-of-words (CBOW) and skip-grams. CBOW model learns its representation from the surrounding words and the skip-gram model learns the context given a word (Figure \ref{nn_lang}).

One way to understand how the training works is by thinking about word embedding training as an autoencoder network \cite{hinton2006reducing} with low-dimensional latent space. The difference is that in the word embedding language model we are learning the representation rather than the weights and weights are often disregarded after the training. The dimensions in this model learn topic directions which are defined by the word co-occurrence. For example, a word "white" is close to a word "black" because they are used in similar context. What is more, this representation enables to perform simple mathematical operations on words such as "\textit{vector("king") - vector("man") + vector("woman") = Queen}".  The biggest problem with this approach was the training time. The model was defined by:

\begin{center}
$O = E \times T \times Q,$
\end{center}
where E is the number of epochs, T is the vocabulary size, and Q is the neural network architecture. Mikolov et al. proposed a more efficient negative sampling method to train the model \cite{mikolov2013distributed}. The idea is, instead of running backpropagation on the entire vocabulary, the training part needs to learn the difference between the real word and some noise sampled from the distribution. This fact led the model to be feasible in the real world applications and the most common choice of current research in NLP. Nevertheless, the model is not perfect. Some possible issues are that the language model does not know how to distinguish between homonyms, word collocations, 
and is often domain specific. This problem continues to be an active area of research.

\begin{figure}[h]
\centering
\includegraphics[width=0.6\textwidth]{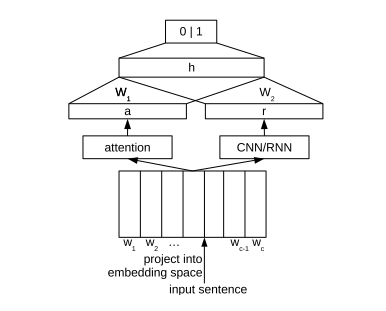}
\caption{Neural network model for uncertainty detection.}
\label{dl_hedge}
\end{figure}

Many recent state-of-the-art algorithms in NLP are achievable with the help of neural network based methods with a combination of word embeddings \cite{cruz2016machine,salas2017feature, conneau2017very}. Adel et al. \cite{adel2017exploring} applied these methods to improve uncertainty detection algorithms that achieve the highest score on the Wikipedia dataset. The model is based on CNN \cite{krizhevsky2012imagenet} and RNN \cite{williams1989learning} with attention mechanism \cite{yang2016stacked}. The model architecture is shown in the figure \ref{dl_hedge}. The main idea is that the attention layer is parallel to the CNN/RNN layer maps the input representation to the output. This method produces 0.67 of F1 score outperforming all other currently available models. 

\begin{figure}[h]
\centering
\includegraphics[width=1\textwidth]{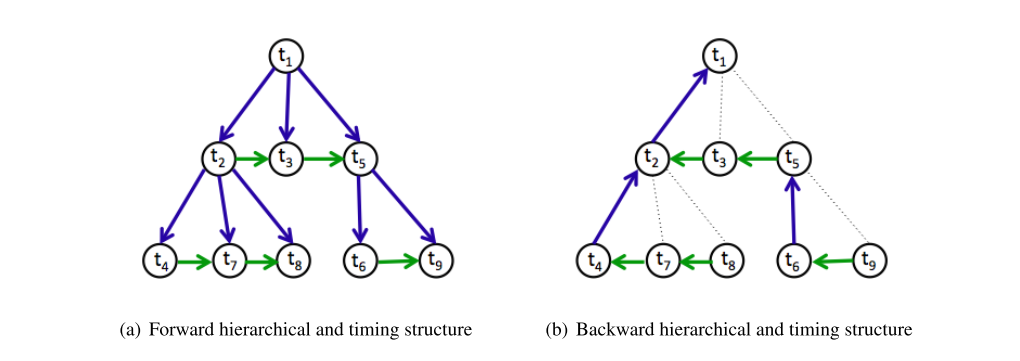}
\caption{Graph-structured LSTM model propagation}
\label{graph_lstm}
\end{figure}

The application of neural networks to the problem of detecting influential users has shown high predictive scores. One of the prominent works in this area is done by V. Zayats and M. Ostendorf \cite{zayats2017conversation}. They proposed a long short-term memory (LSTM) \cite{hochreiter1997long} based approach to predict karma status of Reddit uses. Since Reddit communication can be naturally represented as a graph, the nodes are the text embeddings of the entire post. This representation captures the linguistic representation of each post. However, in the area of SNA, the structural analysis is known to have significant predictive power. For this reason, Zayats et al. presented graph-structured LSTM (Figure \ref{graph_lstm}). The model captures the hierarchical relationship as well as temporal. The evaluation is based on the user karma score that is quantized into 7 classes. 

The biggest downside to neural network based algorithms is that it is hard to understand why the algorithms produce one or another prediction. In other words, when we predict a user to have a high social status in this community, we cannot explain what the prediction is based on. This was the reason why Zayats et al. based their work on n-gram model as the input. For instance, they found that words that represent humor, positive feedback, and emotions are indicative of high karma on Reddit website. This work is also general enough to be applied to other social network websites with a room to improve the performance by using word embeddings instead of n-gram.

\chapter{Future Work and Conclusion} 
\section{Future Work}
\label{fw}
While the problem of hierarchy detection is not new and much work had been done in this domain, it remains a challenge. Researchers often try to solve this problem by directly modeling a user's language or analyzing the network structure to predict relationships among users. However, the real breakthrough will likely to come from a combination of different systems that work together. The early work was based on word analysis and often included parts that predict hedges, politeness, etc. Such systems showed promising results. With the advances of neural networks, similar neural network based methods should help identify high-status users online. 

When word embeddings became usable in real-life applications, this promoted a wave of improvements. Nevertheless, this word representation has many downsides. One of the biggest problems is that the words that are opposite in meaning are often close is the embedded space. One way to improve the model is by post-processing the embedded space. Nikola Mrksic, et. al. \cite{mrkvsic2016counter} proposed a method to counter-fit antonyms and synonyms. This method is shown to improve word representation and potentially will give further improvements in the area of NLP.

The structure-based analysis works well on predicting influential users in online communities. At the same time, the methods haven't changed much in the past years. While a number of graph embedding algorithms have been proposed \cite{niepert2016learning}, in it unclear how they would perform on this task. Graph embedding is a new, unexplored domain in terms of hierarchy detection. An integration of this approach into the pipeline can improve the performance.

The area of NLP produces new discoveries on a daily basis. These discoveries can help to improve multiple subtasks for hierarchy detection. A single system will require a knowledge from different sub-domains of NLP and graph theory to improve current methods of social network analysis.

\section{Conclusion}
\label{conc}
This work presents a survey of current methods on social relationship analysis in social networks. Despite the difficulty of identifying hierarchical relationships, scientists have shown improvements in this area. Many factors can indicate social status. The number of friends, the frequency of interactions, the word choice, politeness, etc. with a combination of current advances in AI are the key to describe who we are communicating with. Nowadays, people are increasing prefer to communicate with their friends in the virtual world where they form communities and establish social connections. Understanding who is who on the Internet is important for many business, politics, and national security. Research in this area will help to understand current trends and human relationships as a whole. 

\backmatter

\bibliographystyle{apalike}
\bibliography{bibl}
\addcontentsline{toc}{chapter}{\numberline{}Bibliography }

\end{document}